# Use of PSO in Parameter Estimation of Robot Dynamics; Part Two: Robustness

Hossein Jahandideh, Mehrzad Namvar

*Abstract*— In this paper, we analyze the robustness of the PSO-based approach to parameter estimation of robot dynamics presented in "Part One". We have made attempts to make the PSO method more robust by experimenting with potential cost functions. The simulated system is a cylindrical robot; through simulation, the robot is excited, samples are taken, error is added to the samples, and the noisy samples are used for estimating the robot parameters through the presented method. Comparisons are made with the least squares, total least squares, and robust least squares methods of estimation.

## I. INTRODUCTION

Inverse dynamics of a robot is obtained in the following form [1]:

$$\tau = D(q)\ddot{q} + C(q,\dot{q})\dot{q} + g(q) + F_c sign(\dot{q}) + F_v \dot{q} \quad (1)$$

where $\tau$ denotes the vector of forces/torques applied to the robot's joints, $D$ is the manipulator inertia matrix, $C$ is the coriolis/centripetal matrix, $g$ is the gravity vector, $F_c$ is the coulomb friction and $F_v$ is the viscous friction. The vector $q$ contains all the joint variables- lengths for prismatic joints, and angles for revolute joints.

Without knowledge of the masses, centers of mass, inertia parameters, and frictional parameters of the robot links, (1) is incomplete. Conventional methods of robot parameter estimation, factorize (1) into the form:

$$\tau = Y(q,\dot{q},\ddot{q})\alpha \quad (2)$$

where $Y$ is the linear regressor and $\alpha$ is the set of base parameters. The base parameters are the minimum number of parameters that influence the dynamic behavior of the robot. They may be combinations of the mass, inertia, friction, and gravity parameters. Finding the set of base parameters is called *parameterization*.

In "part one" [2], particle swarm optimization (PSO) was used to estimate the robot parameters without the step of parameterization. In this paper, we wish to examine the performance of the PSO approach to estimation, on the condition where the samples which are to be used for estimation have relatively large errors. This analysis will be performed by comparison with the least squares (LS), total least squares (TLS), and robust least squares (RLS) methods.

Mehrzad Namvar is a faculty member of the control systems group in the Electrical Engineering department at Sharif University of Technology, Tehran, Iran. He received his PhD in Control systems in 2001 from the Grenoble Institute of Technology (INPG) in France.  namvar@sharif.ir

Hossein Jahandideh is a student in the Electrical Engineering department at Sharif University of Technology, Tehran, Iran (2008-2013) hs.jahan@gmail.com

All methods are simulated for the estimation of the parameters of a cylindrical robot. The cylindrical robot has only four identifiable (influential) parameters. The four parameters are estimated by the four methods and the estimated values are compared to the real parameter values given to the simulated robot for excitation and sampling. The LS, TLS, and RLS methods require parameterization of the inverse dynamics, which we have thus provided and used for this purpose. We have made efforts to make improvements to the PSO approach.

A relevant research to this paper is [3], which has experimented with robust cost functions for the PSO, to identify complex nonlinear systems (not particularly robots). Similarly, in this paper we have experimented with potential cost functions, though not the ones tested in [3]. Unlike [3], in which all compared methods are PSO-based, we have made comparisons to the RLS method to achieve justifiable results.

Previous research on reducing the effects of measurement noise on robot parameter estimation is found in [4] and [5]. The LS-based method in [4] depends on data filtering, and data filtering depends on a very high sampling frequency. Similarly, a weighted least squares (WLS) approach was proposed in [5], which relies on data filtering and knowledge of the exact properties of the noise. On the contrary, the TLS and RLS solutions [6-9] are designed for noisy conditions; they do not rely on data filtering, and thus can be obtained by a limited number of samples (as long as the number of samples exceeds the number of identifiable parameters). Likewise, the PSO does not rely on a high sampling rate; hence in this paper the PSO approach is compared to the TLS and RLS methods rather than methods based on data filtering such as in [4] and [5]. In this paper we take a limited number of samples for our estimation and the properties of noise are not considered.

The rest of this paper is organized as follows: In section 2, PSO is introduced. The least squares methods are introduced in section 3. The simulated experiment (including the introduction of the cylindrical robot) is explained in detail in section 4. In section 5, the results of the simulation are summarized. Conclusions are drawn in section 6.

## II. PARTICLE SWARM OPTIMIZATION

Particle Swarm Optimization (PSO) is a swarm intelligence optimization algorithm inspired by simulating bird flocking or fish schooling. Examples of swarm and evolutionary algorithms and their applications are explained in an orderly manner in [10]. PSO was first introduced by Kennedy and Eberhart in [11]. The mathematical analyses behind PSO were explained by Clerk and Kennedy in [12].

PSO has been utilized in a wide range of scientific fields (including robotics, control, and system identification), examples of which can be found in [3] and [13-15].

Let $f: \Re^n \rightarrow \Re$ be the function to be optimized. Without loss of generality, we'll take our objective to be minimization.

Objective: minimize $f(x)$
subject to: $x \epsilon \chi$

The constraint $x \epsilon \chi$ can be efficiently merged with the function $f(x)$ [20]. PSO algorithm uses a swarm of k particles as agents to search for the optimal solution in an n-dimensional space. The starting position of a particle is randomly set within the range of possible solutions to the problem. The range is determined based on an intuitive guess of the maximum and minimum possible values of each component of x, but doesn't need be accurate. Each particle analyzes the function value ($f(p)$) of its current position (p), and has a memory of its own best experience (Pbest), which is compared to p in each iteration, and is replaced by p if $f(p) < f(Pbest)$. Aside from its own best experience, each particle has knowledge of the best experience achieved by the entire swarm (the global best experience denoted by Gbest). Based on the data each agent has, its movement in the i-th iteration is determined by the following formula:

$$V_i = w_i V_{i-1} + C_1 r_1 (Pbest - p_{i-1}) + C_2 r_2 (Gbest - p_{i-1}) \quad (3)$$

where $V_i$, $P_i$, Pbest, and Gbest are n-vectors (or similar objects, such as matrices with n components), $r_1$ and $r_2$ are random numbers between 0 and 1, re-generated at each iteration. $C_1$ and $C_2$ are constant positive numbers, $C_1$ is the *cognitive learning rate* and $C_2$ is the *social learning rate*. $w_i$ is the *inertia weight*, the importance of which is comprehensively discussed in [16]. The new position of each particle at the i-th iteration is updated by:

$$p_i = p_{i-1} + V_i \quad (4)$$

After certain conditions are met, the iterations stop and the Gbest at the latest iteration is taken as the optimal solution to the problem. In this paper, we let the PSO algorithm end when the number of iterations reaches a certain number.

In [2], we used PSO for estimating robot parameters by using the inverse dynamics function defined in robot simulation software (in our case, [17]).

Each sample we have of the robot dynamics contains the following data: $\tau_{(i)}$, $q_{(i)}$, $\dot{q}_{(i)}$, $\ddot{q}_{(i)}$, where $\tau$ is the n-vector of forces/torques, and q is the state of the n joint variables (n is equal to the degrees of freedom of the robot). The index (i) is used for the i-th sample. For each sample, based on the estimated parameters and the inverse dynamics model, a vector $\hat{\tau}$ can be calculated by the cited software.

Define $E_{(i)}$ for the i-th sample:

$$E_{(i)} = \tau_{(i)} - \hat{\tau}_{(i)} \quad (5)$$

Now define the error matrix E for which the i-th column is $E_{(i)}$:

$$E = [E_{(1)} | E_{(2)} | ... | E_{(N)}] \quad (6)$$

where N is the number of samples available. The cost function for the PSO algorithm is defined for the matrix E; for example (7).

$$f(E) = \|E\|_2 \quad (7)$$

The objective of the PSO algorithm in our estimation task is to find the set of parameters that minimizes the cost function $f$.

In this paper, we wish to improve the performance of the PSO algorithm in terms of robustness. Many contributions have been made to modifying and improving the PSO algorithm (e.g. [18-20]). However, these modifications have been made to the algorithm itself, while in our application, it is seen that due to the sample errors, the cost function of the real parameter values is higher than that of the estimated values. This implies that regardless of the modifications made to the algorithm, the estimated values will not improve unless the cost function is modified. Thus in this paper we have experimented with different candidates for the cost function. The advantage of the PSO algorithm over analytical methods such as LS is the flexibility of the cost function. The cost function may be non-linear, non-convex, and/or non-differentiable. The cost function may even be discontinuous at certain points.

### III. THE LEAST SQUARES METHODS

#### A. Least Squares

The conventional method for robot parameter estimation is the least squares method and its derivatives. [21] is a very old reference which introduces LS as a system identification tool. Examples of research which have based robot parameter estimation on LS are found in [4] and [22-25].

In the LS approach, the inverse dynamics are parameterized as in (2). The regressor $Y$ is computed for the data from each sample to create samples of the regressor. If the sample-regressor for the i-th sample is denoted by $Y_{(i)}$, the *observation matrix* W is defined by:

$$W = \begin{bmatrix} Y_{(1)}^T & Y_{(2)}^T & ... & Y_{(N)}^T \end{bmatrix}^T \quad (8)$$

where N is the number of samples taken. $\tau_s$ is defined by:

$$\tau_s = \begin{bmatrix} \tau_{(1)}^T & \tau_{(2)}^T & ... & \tau_{(N)}^T \end{bmatrix}^T \quad (9)$$

where $\tau_{(i)}$ is an n×1 matrix containing the torques(forces) of the i-th sample and N is the total number of samples. The LS estimation of α minimizes $f_{LS}$ defined by:

$$f_{LS(\alpha)} = \|\tau_s - W\alpha\|_2 \quad (10)$$

It is known from [4, 11, 12, 16, 21-25] that the solution to this problem is given by:

$$\hat{\alpha}_{LS} = (W^T W)^{-1} W^T \tau_s \quad (11)$$

### B. Total Least Squares

The TLS solution [8, 9] ($\hat{\alpha}_{TLS}$) minimizes ρ defined by the following formulae:

$$(W_{LS} + \Delta W)\alpha = \tau_s + \Delta \tau \quad (12)$$
$$\Delta = [\Delta W \mid \Delta \tau] \quad (13)$$
$$\rho = \|\Delta\|_F \quad (14)$$

where the matrices $\Delta W$ and $\Delta \tau$ are defined such that there exists a unique vector of parameter values α that solves (12). $\|.\|_F$ denotes the Frobenius norm of a vector or matrix (in this case a vector).

It has been shown in [9] that the TLS solution can be obtained by using singular value decomposition, which results in the following solution:

$$\hat{\alpha}_{TLS} = (W^T W - \sigma^2 I)^{-1} W^T \tau_s \quad (15)$$

where σ denotes the smallest singular value of the matrix $[W|\tau_s]$. σ is zero if $\tau_s$ is linearly dependant on the columns of W (i.e. if there exists an α that solves $W\alpha = \tau_s$).

### C. Robust Least Squares

The RLS method was introduced in [6, 7]; it is known to be less accurate, while being more robust (i.e. less risky) than the TLS method as it takes into account the worst situation that our measurement data could have.

The RLS solution ($\hat{\alpha}_{RLS}$) minimizes r defined by the following formulae:

$$r = \max \|(W + \Delta W)\alpha - (\tau_s + \Delta \tau)\|_2 \quad (16)$$

Subject to:

$$\|[\Delta W \mid \Delta \tau]\|_F \leq \rho \quad (17)$$

where ρ is the perturbation bound. If ρ is unknown (such as considered in our problem), the ρ obtained from (14) by solving the TLS problem may be used in the RLS problem.

The RLS problem can be formulated as a second-order cone problem [7]:

$$\text{minimize } \lambda \text{ subject to: } \left\| W/\rho \, \alpha - \tau_s/\rho \right\| \leq \lambda - \gamma, \; \left\| \begin{bmatrix} \alpha \\ 1 \end{bmatrix} \right\| \leq \gamma \quad (18)$$

In this paper the solution to this problem is obtained with the help of CVX, a package for specifying and solving convex programs [26, 27].

## IV. SIMULATED EXPERIMENT

The robot used in our simulation is a cylindrical robot. The cylindrical robot is depicted in fig. 1. The link parameters of a cylindrical robot, following the traditional notation, are given in table 1 [22].

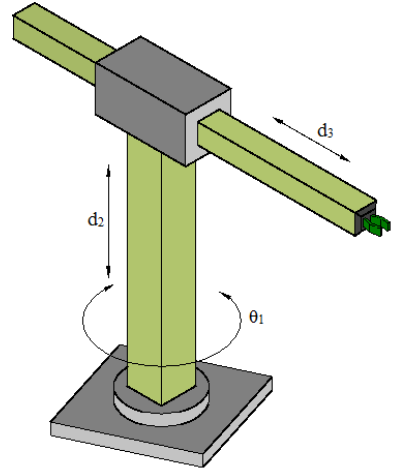

Figure 1. A simple depiction of the cylindrical robot

TABLE I. THE LINK PARAMETERS OF THE CYLINDRICAL ROBOT

| link number (i) | $a_{(i)}$(m) | $\alpha_{(i)}$(rad) | $d_{(i)}$(m) | $q_{(i)}$ |
|---|---|---|---|---|
| 1 | 0 | 0 | 0 | $\theta_1$ |
| 2 | 0 | -π/2 | 0 | $d_2$ |
| 3 | 0 | 0 | 0 | $d_3$ |

The inverse dynamics equations of the cylindrical robot are given by [22]:

$$\tau_1 = [(I_{1zz} + I_{2yy} + I_{3yy}) + m_3(s_{3z} + d_3)^2]\ddot{\theta}_1 \\ + [2m_3(s_{3z} + d_3)]\dot{d}_3\dot{\theta}_1 \quad (19)$$

$$\tau_2 = (m_2 + m_3)(\ddot{d}_2 + g) \quad (20)$$

$$\tau_3 = m_3 \ddot{d}_3 - m_3(s_{3z} + d_3)\dot{\theta}_1^2 \quad (21)$$

where $s_{3z}$ denotes the component of the center of mass of the third link along its own z-axis; $\theta_1$, $d_2$, and $d_3$ are the joint variables, $m_i$ is the mass of the i-th link, g is the gravity force, and $I_{abc}$ is the bc component of the moment of inertia of link a about its center of mass. The frictional factors have been omitted for simplicity.

(19-21) have been derived with the assumption that link 1 is zero dimensional and links 2 and 3 each are one dimensional figures. We have chosen this simple example to be able to easily compare the estimations obtained by the different methods. There are only four parameters to be estimated; $m_2$, $m_3$, $s_{3z}$, and $I$. $I$ is defined as:

$$I = I_{1zz} + I_{2yy} + I_{3yy} \quad (22)$$

The inverse dynamics of the cylindrical robot are parameterized into the form (2) as:

$$Y = \begin{bmatrix} \ddot{\theta}_1 & 0 & 2d_3\dot{d}_3\dot{\theta}_1 + d_3^2\ddot{\theta}_1 & 2\dot{d}_3\dot{\theta}_1 + 2d_3\ddot{\theta}_1 \\ 0 & \ddot{d}_2 + g & \ddot{d}_2 + g & 0 \\ 0 & 0 & \ddot{d}_3 - d_3\dot{\theta}_1^2 & -\dot{\theta}_1^2 \end{bmatrix} \quad (23)$$

$$\alpha = \begin{bmatrix} I_{1zz} + I_{2yy} + I_{3yy} + m_3 s_{3z}^2 \\ m_2 \\ m_3 \\ m_3 s_{3z} \end{bmatrix} \quad (24)$$

Once α is estimated by a least squares method, in order to compare the estimated parameters to those estimated by PSO, the values of the four parameters $m_2$, $m_3$, $s_{3z}$, and $I$ must be extracted from α.

There are countless potential cost functions that can be defined for the PSO algorithm; here we will consider 16 of these. It is possible that replacing the error matrix E (from (6)) by $E_{rel}$ defined by the following equations might yield better results:

$$T = [\tau_{(1)} | \tau_{(2)} | \ldots | \tau_{(N)}] \quad (25)$$

$$\hat{T} = [\hat{\tau}_{(1)} | \hat{\tau}_{(2)} | \ldots | \hat{\tau}_{(N)}] \quad (26)$$

$$E_{rel}: \quad E_{rel\_ij} = (T_{ij} - \hat{T}_{ij}) / T_{ij} \quad (27)$$

The eight cost functions considered are:

$$f_1(E) = \|E\|_F \quad (28)$$

$$f_2(E) = \|E\|_2 \quad (29)$$

$$f_3(E) = \|E\|_1 \quad (30)$$

$$f_4(E) = \|E\|_\infty \quad (31)$$

$$f_5(E) = \|E\|_1 \|E\|_\infty \quad (32)$$

$$f_6(E) = \|E\|_1 \|E\|_2 \|E\|_\infty \quad (33)$$

$$f_7(E) = \max_i (\sum_j E_{ij}^2) \quad (34)$$

$$f_8(E) = \max_{i,j}(E_{ij}) \quad (35)$$

The same eight cost functions are defined for $E_{rel}$ as well ($f_9$ to $f_{16}$ respectively). It is notable that the CVX can be used to solve the LS problem for the relative errors, substituting the relative error vector for the absolute error vector in (10). TLS and RLS can also be defined for relative errors by replacing W and $\tau_s$ by:

$$Wr: \quad Wr_{ij} = W_{ij} / \tau_{s(i)} \quad (36)$$

$$\tau_r: \quad \tau_{r(i)} = 1 \quad (37)$$

Hence, we have also obtained the relative LS, TLS, and RLS solutions in our experiments, denoted respectively by LS-rel, TLS-rel, and RLS-rel in tables 2 to 5.

It may be thought that since the 2-norm, 1-norm, and infinity-norm are convex functions, CVX can be used to obtain the optimal values for f2, f3, and f4 (f10, f11, and f12). This is not true, due to that the LS methods use the error vector, whereas the PSO methods use the error matrix. If the error matrix is defined as a function of the error vector, and f2, f3, or f4 (f10, f11, or 12) are used on the resulting matrix, the total function is a non-convex function of the parameters in α.

For the PSO algorithm, the learning rates of $C_1$ and $C_2$ in (3) are both set to 1.3, and $w_i$ decreases linearly from 0.9 to 0.4 through the first 100 iterations and is set at 0.4 for the next 200 iterations (the total number of iterations are set at 300). The swarm population is set to 20.

The reference trajectory used for sampling is planned according to the PSO-based method presented in [2]. Samples are taken, random error is given to the samples, and the same samples are used for all 11 methods of estimation (i.e. the LS, TLS, RLS, and the 8 PSO methods); the results are compared. The physical boundaries observed by the planned trajectory are as follows:

$$-\pi \le \theta_1(rad) \le \pi, \quad -4 \le \dot{\theta}_1(\tfrac{rad}{s}) \le 4, \quad -3 \le \ddot{\theta}_1(\tfrac{rad}{s}) \le 3$$

$$0 \le d_2(m) \le 1, \quad -2 \le \dot{d}_2(\tfrac{m}{s}) \le 2, \quad -2 \le \ddot{d}_2(\tfrac{m}{s}) \le 2 \quad (38)$$

$$0 \le d_3(m) \le 1, \quad -1.5 \le \dot{d}_3(\tfrac{m}{s}) \le 1.5, \quad -1 \le \ddot{d}_3(\tfrac{m}{s^2}) \le 1$$

The PSO-planned trajectory is as follows:

$$\theta_{1(t)} = 0.43\sin(2.2t) + 0.23\sin(1.8t) - 3.4\sin(0.06t) - 0.36$$

$$d_{2(t)} = \sin(0.1t) - 0.3\sin(0.07t) + 0.35\sin(1.3t) - 0.014$$

$$d_{3(t)} = 0.1\sin(0.1t) - 0.1\sin(2.7t) + 0.06\sin(0.14t) + 0.26$$

$$0 \le t \le 10 \quad (39)$$

Tables 2-5 show the estimated parameters based on different sets of samples. In table 2, ten samples are used; each state measurement (i.e. $q$, $\dot{q}$, $\ddot{q}$) of all samples is given a random error of up to 20%. In table 3, 10 samples are used and errors of up to 20% are placed upon on all force/torque measurements (i.e. $\tau$). Table 4 shows the estimated values based on 20 samples with errors of up to 20% on all measurements (both state measurements and force/torque measurements). Table 5 is based on samples which have up to 70% error on ten sample components (of total 120 sample components, i.e. 10(samples)×12(states and force/torques)) and up to 5% error on the rest. The values stated as the PSO estimates are the mean estimate of ten PSO runs, omitting all estimates with cost functions that are relatively too large. The real values stated in the tables are the values dictated to the simulated robot from which the samples are obtained.

## V. DATA ANALYSIS

It is seen in table 2 that for the case where the measurement errors are on the state samples only, using the absolute error matrix (vector, for the LS methods) rather than the relative error matrix (vector) yields more accurate results. However, in table 3, where the errors are on the force/torque measurements only, table 4, where the errors are on all data, and table 5, where few, very large errors exist, for most methods using relative errors seems to result in more accurate estimations. Thus we conclude that using relative errors is more reliable unless it is known that only the state measurements have notable errors, and that all errors are relatively small.

TABLE II. ESTIMATED VALUES BASED ON THE FIRST SET OF SAMPLES

|  | $m_2$ | $m_3$ | $-s_{3z}$ | I | computation time |
|---|---|---|---|---|---|
| real values | 5 | 3 | 0.5 | 3 |  |
| LS | 5.44 | 2.56 | 0.52 | 2.78 | 2 μs |
| TLS | 5.45 | 2.54 | 0.53 | 2.78 | 4 μs |
| RLS | 5.42 | 2.58 | 0.52 | 2.78 | 3.23 s |
| LS-rel | 5.52 | 2.48 | 0.58 | 2.56 | 3 μs |
| TLS-rel | 5.74 | 2.47 | 0.58 | 2.58 | 5 μs |
| RLS-rel | 5.31 | 2.48 | 0.58 | 2.54 | 3.21 s |
| PSO-$f_1$ | 5.43 | 2.57 | 0.52 | 2.78 | 3.21 s |
| PSO-$f_2$ | 5.3 | 2.67 | 0.51 | 2.79 | 3.35 s |
| PSO-$f_3$ | 5.39 | 2.6 | 0.5 | 2.82 | 3.12 s |
| PSO-$f_4$ | 5.06 | 2.93 | 0.51 | 2.8 | 3.00 s |
| PSO-$f_5$ | 4.83 | 3.17 | 0.46 | 2.78 | 3.43 s |
| PSO-$f_6$ | 5.4 | 2.61 | 0.51 | 2.78 | 3.69 s |
| PSO-$f_7$ | 5.1 | 2.9 | 0.5 | 2.79 | 3.77 s |
| PSO-$f_8$ | 4.85 | 3.11 | 0.49 | 2.78 | 3.37 s |
| PSO-$f_9$ | 5.33 | 2.48 | 0.59 | 2.66 | 3.42 s |
| PSO-$f_{10}$ | 5.53 | 2.49 | 0.59 | 2.66 | 3.15 s |
| PSO-$f_{11}$ | 3.84 | 2.72 | 0.58 | 3.07 | 3.11 s |
| PSO-$f_{12}$ | 5.08 | 2.4 | 0.59 | 2.63 | 3.55 s |
| PSO-$f_{13}$ | 4.68 | 2.54 | 0.58 | 2.55 | 3.19 s |
| PSO-$f_{14}$ | 5.09 | 2.56 | 0.58 | 2.62 | 3.68 s |
| PSO-$f_{15}$ | 4.65 | 2.54 | 0.59 | 2.62 | 3.98 s |
| PSO-$f_{16}$ | 4.41 | 2.46 | 0.59 | 2.61 | 3.37 s |

TABLE III. ESTIMATED VALUES BASED ON THE SECOND SET OF SAMPLES

|  | $m_2$ | $m_3$ | $-s_{3z}$ | I | computation time |
|---|---|---|---|---|---|
| real values | 5 | 3 | 0.5 | 3 |  |
| LS | 5.23 | 3.33 | 0.52 | 3.28 | 2 μs |
| TLS | 6.03 | 2.54 | 0.87 | 2.86 | 4 μs |
| RLS | 4.99 | 3.57 | 0.35 | 3.23 | 3.04 s |
| LS-rel | 5.27 | 3.27 | 0.5 | 3.29 | 3 μs |
| TLS-rel | 5.31 | 3.27 | 0.5 | 3.3 | 5 μs |
| RLS-rel | 5.22 | 3.27 | 0.5 | 3.28 | 3.18 s |
| PSO-$f_1$ | 5.21 | 3.34 | 0.5 | 3.32 | 3.27 s |
| PSO-$f_2$ | 4.55 | 4.00 | 0.84 | 1.96 | 3.21 s |
| PSO-$f_3$ | 4.25 | 4.59 | 0.55 | 2.95 | 2.99 s |
| PSO-$f_4$ | 4.78 | 3.66 | 0.81 | 2.47 | 3.12 s |
| PSO-$f_5$ | 4.21 | 4.36 | 0.18 | 3.11 | 3.58 s |
| PSO-$f_6$ | 5.02 | 3.55 | 0.47 | 3.1 | 3.77 s |
| PSO-$f_7$ | 4.77 | 3.78 | 0.53 | 3.14 | 4.02 s |
| PSO-$f_8$ | 4.34 | 4.44 | 0.57 | 3.36 | 3.33 s |
| PSO-$f_9$ | 4.98 | 3.31 | 0.5 | 3.31 | 3.15 s |
| PSO-$f_{10}$ | 4.63 | 3.37 | 0.5 | 3.3 | 3.15 s |
| PSO-$f_{11}$ | 5.05 | 3.38 | 0.49 | 3.16 | 3.37 s |
| PSO-$f_{12}$ | 4.95 | 3.41 | 0.5 | 3.19 | 3.41 s |
| PSO-$f_{13}$ | 5.17 | 3.33 | 0.5 | 3.19 | 3.55 s |
| PSO-$f_{14}$ | 4.64 | 3.4 | 0.49 | 3.35 | 3.61 s |
| PSO-$f_{15}$ | 5.29 | 3.26 | 0.5 | 3.32 | 3.91 s |
| PSO-$f_{16}$ | 5.31 | 3.31 | 0.5 | 3.25 | 3.41 s |

TABLE IV: ESTIMATED VALUES BASED ON THE THIRD SET OF SAMPLES

|  | $m_2$ | $m_3$ | $-s_{3z}$ | I | computation time |
|---|---|---|---|---|---|
| real values | 5 | 3 | 0.5 | 3 |  |
| LS | 6.11 | 2.65 | 0.52 | 3.01 | 5 μs |
| TLS | 20.86 | -12.03 | 0.38 | 3.14 | 9 μs |
| RLS | 5.38 | 3.37 | 0.3 | 2.84 | 6.88 s |
| LS-rel | 6.08 | 2.64 | 0.54 | 3.09 | 8 μs |
| TLS-rel | 6.41 | 2.63 | 0.54 | 3.11 | 12 μs |
| RLS-rel | 5.77 | 2.64 | 0.54 | 3.06 | 6.93 s |
| PSO-$f_1$ | 6.1 | 2.65 | 0.52 | 3.01 | 6.88 s |
| PSO-$f_2$ | 5.69 | 3.07 | 0.51 | 2.2 | 7.02 s |
| PSO-$f_3$ | 5.23 | 3.52 | 0.56 | 2.5 | 7.11 s |
| PSO-$f_4$ | 4.24 | 4.3 | 0.49 | 3.74 | 6.95 s |
| PSO-$f_5$ | 4.52 | 4.24 | 0.56 | 2.55 | 7.22 s |
| PSO-$f_6$ | 5.08 | 3.68 | 0.6 | 2.39 | 7.74 s |
| PSO-$f_7$ | 4.63 | 4.13 | 0.33 | 3.46 | 8.32 s |
| PSO-$f_8$ | 4.63 | 4.14 | 0.64 | 2.6 | 7.56 s |
| PSO-$f_9$ | 6.04 | 2.64 | 0.55 | 3.17 | 7.07 s |
| PSO-$f_{10}$ | 5.1 | 2.68 | 0.54 | 3.02 | 7.23 s |
| PSO-$f_{11}$ | 3.44 | 2.59 | 0.54 | 2.96 | 7.14 s |
| PSO-$f_{12}$ | 5.38 | 2.66 | 0.55 | 3.13 | 7.43 s |
| PSO-$f_{13}$ | 5.09 | 2.68 | 0.54 | 3.06 | 7.91 s |
| PSO-$f_{14}$ | 5.63 | 2.61 | 0.54 | 2.98 | 8.06 s |
| PSO-$f_{15}$ | 4.54 | 2.65 | 0.54 | 3.23 | 8.16 s |
| PSO-$f_{16}$ | 4.31 | 2.59 | 0.54 | 3.35 | 6.98 s |

TABLE V: ESTIMATED VALUES BASED ON THE FOURTH SET OF SAMPLES

|  | $m_2$ | $m_3$ | $-s_{3z}$ | I | computation time |
|---|---|---|---|---|---|
| real values | 5 | 3 | 0.5 | 3 |  |
| LS | 5.42 | 2.73 | 0.5 | 3.12 | 5 μs |
| TLS | 8.52 | -0.35 | -2.57 | 6.3 | 10 μs |
| RLS | 4.92 | 3.22 | 0.32 | 3.04 | 7.15 s |
| LS-rel | 5.5 | 2.61 | 0.52 | 2.98 | 8 μs |
| TLS-rel | 6.06 | 2.59 | 0.52 | 3.03 | 14 μs |
| RLS-rel | 5.00 | 2.62 | 0.52 | 2.92 | 7.22 s |
| PSO-$f_1$ | 5.41 | 2.74 | 0.5 | 3.12 | 6.88 s |
| PSO-$f_2$ | 4.91 | 3.23 | 0.65 | 2.67 | 7.13 s |
| PSO-$f_3$ | 4.48 | 3.39 | 0.52 | 3.00 | 6.79 s |
| PSO-$f_4$ | 6.03 | 2.15 | 0.54 | 3.09 | 7.34 s |
| PSO-$f_5$ | 5.43 | 2.74 | 0.56 | 3.07 | 7.26 s |
| PSO-$f_6$ | 4.85 | 3.32 | 0.43 | 3.07 | 7.64 s |
| PSO-$f_7$ | 4.69 | 3.46 | 0.6 | 3.45 | 7.98 s |
| PSO-$f_8$ | 4.58 | 3.3 | 0.81 | 2.7 | 7.63 s |
| PSO-$f_9$ | 5.39 | 2.64 | 0.52 | 2.98 | 7.09 s |
| PSO-$f_{10}$ | 5.16 | 2.65 | 0.5 | 2.42 | 7.42 s |
| PSO-$f_{11}$ | 6.01 | 2.15 | 0.54 | 2.88 | 7.15 s |
| PSO-$f_{12}$ | 4.34 | 2.95 | 0.51 | 3.1 | 7.25 s |
| PSO-$f_{13}$ | 4.49 | 2.94 | 0.51 | 3.06 | 8.01 s |
| PSO-$f_{14}$ | 5.43 | 2.73 | 0.49 | 3 | 8.32 s |
| PSO-$f_{15}$ | 3.46 | 2.62 | 0.5 | 3.28 | 8.18 s |
| PSO-$f_{16}$ | 3.53 | 2.1 | 0.51 | 2.4 | 7.08 s |

The TLS method is designed to be the most accurate in the case where the errors are very small and almost negligible. As is seen in tables 2 to 5, TLS is an unstable method when the errors grow; tables 4 and 5 are especially notable for their original TLS estimation being completely off-limits. The RLS estimation, both the original and "rel" versions, are seen to give better results than their LS counterparts in nearly all cases. It is predictable and verified that PSO-f1 and PSO-f9 would result in approximately the same estimates as LS and LS-rel respectively.

In table 2, many of the PSO methods have resulted in better estimations than RLS methods, and in each table, there are one or more PSO estimates which surpass the RLS-estimate in terms of accuracy. However, as it is uncertain which case of errors corresponds to our data, we must choose a PSO cost function that surpasses the RLS-rel estimate in almost all cases. f13 seems to be the function we seek. f13 considers both the infinity-norm and the one-norm, and was thus hypothesized and verified to be a robust method. f13 is a non-convex function, thus analytical methods based on convex optimization cannot be used to minimize the function (particularly with the help of the CVX software); this is one of the advantages of the PSO approach. The 2-norm by itself (PSO-f10) does not seem to be a robust method (see tables 2 and 4 where the result of RLS-rel greatly surpasses that of PSO-10) however, adding the 2-norm to f13 to build f14, in some cases seems to cancel out the deficiencies of f13. An average of the results of PSO-f13 and PSO-f14 can be taken as the final estimation, which is seen to be more accurate than the RLS-rel method in all four cases of error type. All the ideas implemented were made possible thanks to the flexibility of PSO, which is another advantage of the PSO approach.

## VI. CONCLUSION

In this paper, we have analyzed the performance of the nonlinear particle swarm optimization (PSO) approach presented in the previous paper, in terms of robustness toward errors on measurement samples. 8 PSO methods (differed by their cost functions) as well as the least squares (LS), total least squares (TLS) and robust least squares (RLS) methods were used to estimate the parameters of a simulated cylindrical robot for 4 types of error conditions. Both the relative force/torque errors and the absolute force/torque errors were considered and compared. It was seen that the consideration of the relative errors, rather than the absolute errors, yield more reliable estimations. The estimation results show that aside from the simplicity of the implementation of the PSO method as described in [2], the PSO approach has the advantage of being flexible and able to minimize non-convex cost functions, which allows the user to tune the PSO method to make it more robust than the RLS method. In particular, it was seen that f13, which mutually considers the 1-norm and the infinite-norm of the relative error matrix, is a robust method, made more robust by using it in conjunction with f-14, which considers also the 2-norm of the relative error matrix. There is much space for experimenting and changing the PSO to make it more robust, experimenting with ideas which have not been implemented in this paper.